\documentclass[sigconf]{acmart}

\pdfoutput=1
\makeatletter
\def\mdseries@tt{m}
\makeatother 

\usepackage{graphicx}
\usepackage{amsmath, amsfonts, amssymb, amsthm} 
\usepackage{hyperref}           
\usepackage{tikz}
\usetikzlibrary{arrows,petri,topaths,backgrounds,patterns,positioning}
\usepackage{pgfplots}
\usepgfplotslibrary{groupplots}
\usepackage{units}
\usepackage{booktabs}
\usepackage{tabularx}
\usepackage{algorithmic}
\usepackage[linesnumbered,lined,boxed]{algorithm2e}
\usepackage[caption=false]{subfig}
\usepgfplotslibrary{fillbetween}
\newsavebox{\mintedbox}
\usepackage{xcolor}
\usepackage{float}
\usepackage{algorithmic}
\usepackage[linesnumbered,lined,boxed]{algorithm2e}
\usepackage{tabularx} 
\usepackage{booktabs}
\usepackage{multirow}
\usepackage[draft=true]{minted}
\usepackage{mathtools}
\usepackage{setspace}
\usepackage{placeins}
\usepackage{tcolorbox}
\usepackage{xparse}
\tcbuselibrary{minted,breakable,xparse,skins}
\usepackage{listings}
\definecolor{bg}{gray}{0.95}
\usepackage{minted}

\AtBeginDocument{%
	\providecommand\BibTeX{{%
			\normalfont B\kern-0.5em{\scshape i\kern-0.25em b}\kern-0.8em\TeX}}}

\newcommand{\specialcell}[2][c]{%
	\begin{tabular}[#1]{@{}c@{}}#2\end{tabular}}

\copyrightyear{2020}
\acmYear{2020}
\setcopyright{acmlicensed}\acmConference[CIKM '20]{Proceedings of the 29th ACM International Conference on Information and Knowledge Management}{October 19--23, 2020}{Virtual Event, Ireland}
\acmBooktitle{Proceedings of the 29th ACM International Conference on Information and Knowledge Management (CIKM '20), October 19--23, 2020, Virtual Event, Ireland}
\acmPrice{15.00}
\acmDOI{10.1145/3340531.3412757}
\acmISBN{978-1-4503-6859-9/20/10}

\begin{document}

\title{Karate Club: An API Oriented Open-Source Python Framework for Unsupervised Learning on Graphs}

\author{Benedek Rozemberczki}
\affiliation{%
  \institution{The University of Edinburgh}
  \city{Edinburgh}
  \country{United Kingdom}}
\email{benedek.rozemberczki@ed.ac.uk}

\author{Oliver Kiss}
\affiliation{%
  \institution{Central European University}
  \city{Budapest}
  \country{Hungary}}
\email{kiss\_oliver@phd.ceu.edu}

\author{Rik Sarkar}
\affiliation{%
  \institution{The University of Edinburgh}
  \city{Edinburgh}
  \country{United Kingdom}}
\email{rsarkar@inf.ed.ac.uk}

\begin{abstract}
Graphs encode important structural properties of complex systems. Machine learning on graphs has therefore emerged as an important technique in research and applications. We present \textit{Karate Club} -- a Python framework combining more than 30 state-of-the-art graph mining algorithms. These unsupervised techniques make it easy to identify and represent common graph features. The primary goal of the package is to make community detection, node and whole graph embedding available to a wide audience of machine learning researchers and practitioners. 
\textit{Karate Club} is designed with an emphasis on a consistent application interface, scalability, ease of use, sensible out of the box model behaviour,  standardized dataset ingestion, and output generation. This paper discusses the design principles behind the framework with practical examples. We show \textit{Karate Club}'s efficiency in learning performance on a wide range of real world clustering problems and classification tasks along with supporting evidence of its competitive speed. 
\end{abstract}


\maketitle

\section{Introduction}\label{sec:karate_club_introduction}

Techniques that extract features from graph data in an unsupervised manner~\cite{deepwalk, bigclam, graph2vec} have seen an increasing success in the machine learning community. Features automatically extracted by these methods can serve as inputs for link prediction, node and graph classification, community detection and various other tasks tasks~~\cite{ musae,deepwalk, graph2vec, walklets, yanardag} in a wide range of real world research and application scenarios. Graph mining tools such as ~\cite{snap, networkx, graphtool} have contributed to enhancement and development of machine learning pipelines. The need for complicated custom feature engineering is reduced by unsupervised graph mining techniques. This approach produces features that are naturally reusable on multiple types of tasks. Recent research~\cite{deepwalk, walklets, netlsd} has made such feature extraction highly efficient and parallelizable.  

The democratization of machine learning for tabular data was led by frameworks which made fast paced development possible. Tools such as~\cite{scikit, scikitapi, tensorflow, pytorch, littleballoffur} are available in general purpose scripting languages with easy to use consistent interfaces. On the other hand, current graph based learning frameworks are less mature and of limited scope~\cite{networkx, graphtool}. For example, these packages host certain community detection algorithms, but none for whole graph or node embedding. In addition, some of these tools~\cite{snap, graphtool} have significant barriers for the end users in terms of installing prerequisites and custom data structures used for representing graphs. 

\noindent \textbf{Present work.} We propose \textit{Karate Club}, an open source Python framework for unsupervised learning on graphs. We implemented \textit{Karate Club} with consistent API oriented design principles in mind which makes the library end user friendly and modular. The name of the package is inspired by Zachary's Karate Club \cite{zachary} -- a network commonly used to demonstrate network algorithms. The design of this machine learning toolbox was motivated by the principles used to create the widely used \textit{scikit-learn} package \cite{scikitapi}. 

The design entails a number of fundamental engineering patterns. Each algorithm has a sensible default hyperparameter setting which helps non expert practitioners. To further ease the use of our package, algorithms have a limited number of shared, publicly available methods (e.g. fit). Models ingest data structures from the scientific Python ecosystem \cite{networkx, scipy, numpy} as input and the generated output also follows these formats. The inner model mechanics are always implemented as private methods using optimized back-end libraries \cite{numpy, scipy, pygsp, gensim} for computing. These principles combined with the extensive documentation ensure that \textit{Karate Club} is accessible to a wider audience than the currently available open-source graph mining frameworks.

Our empirical evaluation focuses on three common graph mining tasks: non-overlapping community detection, node and graph classification. We compare the learning performance of node and graph level algorithms implemented in our framework on various real world social, web and collaboration networks (collected from Deezer, Reddit, Facebook, Twitch, Wikipedia and GitHub). With respect to the runtime, models in \textit{Karate Club} show reasonable scalability which we demonstrate by the use of synthetic data.
\clearpage

\noindent\textbf{Our contributions.} Specifically our work makes the following key contributions:
\begin{enumerate}
	\item We release \textit{Karate Club}, a Python graph mining framework which provides a wide range of easy to use community detection, node and whole graph embedding procedures.
	\item We demonstrate with code the main ideas of the API oriented framework design: hyperparameter encapsulation and inspection, available public methods, dataset ingestion, output generation, and interfacing with downstream learning algorithms and evaluation methods.
	\item We evaluate the learning performance of the framework on real world community detection, node and graph classification problems. We validate the scalability of the algorithms implemented in our framework. 
	\item We open sourced with the framework a detailed documentation and released multiple large graph classification datasets.
\end{enumerate}

The remainder of this paper is structured as follows. In Section \ref{sec:procedures} we discuss the covered graph mining techniques. We overview the main principles behind \textit{Karate Club} in Section \ref{sec:karate_club_design} where we included detailed examples to explain these design ideas. The learning performance and scalability of the algorithms in the package are evaluated in Section \ref{sec:karate_club_experiments}. We review related data mining libraries in Section \ref{sec:karate_club_related_work}. The paper concludes with Section \ref{sec:karate_club_conclusion} where we discuss future directions. The source code of \textit{Karate Club} can be downloaded from \url{https://github.com/benedekrozemberczki/karateclub}; the Python package can be installed from the \textit{Python Package Index}. Extensive documentation is available at 
\url{https://karateclub.readthedocs.io/en/latest/} with detailed examples.
\section{Graph mining procedures in {\em Karate Club}}\label{sec:procedures}

In this section, we briefly discuss the various graph mining algorithms which are available in the 1.0. release of the \textit{Karate Club} package. 

\subsection{Community detection}
Community detection techniques cluster the vertices of the graph into densely connected groups of nodes. This grouping can be the final result or an input for a downstream learning algorithm (e.g. node classification or link prediction).

{\em Karate Club} currently contains several methods for overlapping and non-overlapping community detection. Non-overlapping community detection is analogous to clustering, and assumes that a node can only belong to a single group; see, for example, \cite{edmot,label_propagation,scd, gemsec}. While overlapping community detection is analogous to fuzzy clustering as nodes have an affiliation with multiple clusters; look for example  \cite{bigclam, danmf, mnmf, nnsed, symmnmf}. 
\subsection{Node embedding}
Node embeddings map vertices of a graph into an Euclidean space in which those that are deemed to be similar according to a certain notion will be in close proximity. The Euclidean representation makes it easier to apply standard machine learning libraries for node classification, link prediction, clustering etc.

\textit{Neighbourhood preserving embedding} maintains the proximity of nodes in the graph when an embedding is created. These methods implicitly \cite{deepwalk, walklets, diff2vec} or explicitly  \cite{grarep,boostne,nmfadmm, netmf} decompose a proximity matrix (e.g. powers of the adjacency matrix or a sum of these matrices) to create the node embedding.

\textit{Structural embedding} conserves the structural roles of nodes in the embedding space \cite{rolx, role2vec, graphwave}. Nodes with similar embeddings have a similar distribution of descriptive statistics (e.g. centrality and clustering coefficient) in their vicinity. Embeddings are distilled by the decomposition of matrices representing structural features of nodes. 

\textit{Attributed embedding} retains the neighbourhood structure and generic feature similarity of nodes when the embedding is learned. This learning involves the joint factorization of the node-node and node-feature matrices with a direct \cite{bane, tadw} or implicit matrix decomposition technique \cite{musae, sine}.

\textit{Meta embedding} combines information from neighbourhood preserving, structural and attributed embeddings in order to create higher representation quality embeddings \cite{neu}.

\subsection{Whole graph embedding and summarization}
Whole graph embedding and summarization techniques create fixed size representations of entire graphs as points in a Euclidean space. Those graphs which are close in the embedding space share structural patterns such as subtrees. These representations are used for a range of graph level tasks -- graph classification, regression and whole graph clustering. 

\textit{Spectral fingerprints} extract statistics from the eigenvectors and eigenvalues of the graph Laplacian \cite{netlsd, sf, fgsd}. Vectors of the descriptive statistics are used as the whole graph representation.

\textit{Implicit factorization} techniques create a graph -- structural feature matrix \cite{graph2vec, gl2vec} by enumerating string features in the graphs. This matrix is decomposed in order to create whole graph descriptors and feature embeddings jointly.
\section{Design Principles}\label{sec:karate_club_design}
When we created \textit{Karate Club}, we used an API oriented machine learning system design point of view \cite{scikit,scikitapi} in order to make an end-user friendly machine learning tool. This API oriented design principle entails a few simple ideas. In this section we discuss these ideas and their apparent advantages with appropriate illustrative examples in great detail.
\subsection{Encapsulated model hyperparameters and inspection}
An unsupervised \textit{Karate Club} model instance is created by using the constructor of the appropriate Python object. This constructor has a \textit{default hyperparameter setting} which allows for sensible out-of-the-box model usage. In simple terms this means that the end user does not need to understand the inner model mechanics in great detail to use the methods implemented in our framework. We set these default hyperparameters to provide a reasonable learning and runtime performance. If needed, these model hyperparameters can be modified at the model instance creation time with the appropriate re-parametrization of the constructor. The hyperparameters are stored as \textit{public attributes} to allow the inspection of model settings.

We demonstrate the encapsulation of hyperparameters by the code snippet in Figure \ref{fig:deepwalk_base}. First, we want to create an embedding for a \textit{NetworkX} generated Erdos-Renyi graph (line 4) with the standard hyperparameter settings. When the model is constructed and fitted (lines 6-7) we do not change default hyperparameters and we can print the standard setting of the dimensions hyperparameter (line 8). Second, we decided to set a different number of dimensions, so we created and fitted a new model (lines 10-11) and we print the new value of the dimensions hyperparameter (line 12).    
\begin{figure}[h!]
\begin{minted}[linenos,fontsize=\small,xleftmargin=0.5cm,numbersep=3pt,frame=lines]{python}
import networkx as nx
from karateclub import DeepWalk
   
graph = nx.gnm_random_graph(100, 1000)

model = DeepWalk()
model.fit(graph)
print(model.dimensions)

model = DeepWalk(dimensions=64)
model.fit(graph)
print(model.dimensions)
\end{minted}
\caption{Creating a synthetic graph, using a DeepWalk model with standard and modified hyperparameter settings.}\label{fig:deepwalk_base}
\end{figure}

\subsection{API Consistency and non-proliferation of classes}
Each unsupervised machine learning model in \textit{Karate Club} is implemented as a separate class which inherits from the \textit{Estimator} class. Algorithms implemented in our framework have a limited number of \textit{public methods} as we do not assume that the end user is particularly interested in the algorithmic details related to a specific technique. All models are trained by the use of the \textit{fit} method which takes the inputs (graph, node features) and calls the appropriate private methods to learn an embedding or clustering. Node and graph embeddings are returned by the \textit{get\_embedding} public method and cluster memberships are retrieved by calling \textit{get\_memberships}.

\begin{figure}[h!]
\begin{minted}[linenos,fontsize=\small,xleftmargin=0.5cm,numbersep=3pt,frame=lines]{python}
import networkx as nx
from karateclub import DeepWalk
   
graph = nx.gnm_random_graph(100, 1000)

model = DeepWalk()
model.fit(graph)
embedding = model.get_embedding()
\end{minted}
\caption{Creating a synthetic graph, using the DeepWalk constructor, fitting the embedding and returning it.}\label{fig:deepwalk}
\end{figure}

We avoided the proliferation of classes with two specific strategies. First, the inputs used by our framework and the outputs generated do not rely on custom data classes. This helps to prevent the unnecessary growth of the number of classes and also helps with interfacing with downstream applications. Second, algorithms which use the same data pre-processing or algorithmic step (e.g. truncated random walk, Weisfeiler-Lehman hashing) were built on shared blocks.

In Figure \ref{fig:deepwalk} we create a random graph (line 4), and a DeepWalk model with the default hyperparameters (line 6), we fit this model (line 7) using the public \textit{fit} method (line 7) and return the embedding by calling the public \textit{get\_embedding} method (line 8).

\begin{figure}[h!]
\begin{minted}[linenos,fontsize=\small,xleftmargin=0.5cm,numbersep=3pt,frame=lines]{python}
import networkx as nx
from karateclub import Walklets
    
graph = nx.gnm_random_graph(100, 1000)

model = Walklets()
model.fit(graph)
embedding = model.get_embedding()
\end{minted}
\caption{Creating a synthetic graph, using the Walklets constructor, fitting the embedding and returning it.}\label{fig:walklets}
\end{figure}

The example in Figure \ref{fig:deepwalk} can be modified to create a \textit{Walklets} embedding with minimal effort by changing the model import (line 2) and the constructor (line 6) -- these modifications result in the snippet of Figure \ref{fig:walklets}.

Looking at these two snippets the advantage of the API driven design is evident as we only needed to do a few modifications. First, we had to change the import of the embedding model. Second, we needed to modify the model construction and the default hyperparameters
were already set. Third, the public methods provided by the \textit{DeepWalk} and \textit{Walklets} classes behave the same way. An embedding is learned with \textit{fit} and it is returned by
\textit{get\_embedding}. This allows for quick and minimal changes to the code when an upstream unsupervised model used for feature extraction performs poorly.

\subsection{Standardized dataset ingestion}\label{subsec:karate_input}

We designed \textit{Karate Club} to use standardized dataset ingestion when a model is fitted. Practically this means that algorithms which have the same purpose use the same data types for model training. In detail: 
\begin{itemize}
	\item Neighbourhood based and structural node embedding techniques use a single \textit{NetworkX} graph as input for the fit method. 
	\item Attributed node embedding procedures take a \textit{NetworkX} graph as input and the features are represented as a \textit{NumPy} array or as a \textit{SciPy} sparse matrix. In these matrices rows correspond to nodes and columns to features.
	\item Graph level embedding methods and statistical graph fingerprints take a list of \textit{NetworkX} graphs as an input. 
	\item Community detection methods use a \textit{NetworkX} graph as an input.
\end{itemize} 

\subsection{High performance model mechanics}\label{subsec:internals}

The underlying mechanics of the graph mining algorithms were implemented using widely available Python libraries which are not operation system dependent and do not require the presence of other external libraries like \textit{TensorFlow} or \textit{PyTorch} does \cite{tensorflow, pytorch}. The internal graph representations in \textit{Karate Club} use \textit{NetworkX}. Dense linear algebra operations are done with \textit{NumPy} and their sparse counterparts use \textit{SciPy}. Implicit matrix factorization techniques \cite{deepwalk,walklets,musae, role2vec, sine} utilize the \textit{GenSim} \cite{gensim} package and methods which rely on graph signal processing use \textit{PyGSP} \cite{pygsp}.

\begin{figure}[h!]
\begin{minted}[linenos,fontsize=\small,xleftmargin=0.5cm,numbersep=3pt,frame=lines]{python}
import community
import networkx as nx
from karateclub import LabelPropagation, SCD

graph = nx.gnm_random_graph(100, 1000)

model = SCD()
model.fit(graph)
scd_memberships = model.get_memberships()

model = LabelPropagation()
model.fit(graph)
lp_memberships = model.get_memberships()

print(community.modularity(scd_memberships, graph))
print(community.modularity(lp_memberships, graph))
\end{minted}
\caption{Creating a synthetic graph, clustering with two community detection techniques and using an external library to evaluate the modularity of clusterings.}\label{fig:integrated}
\vspace{-5mm}
\end{figure}
\subsection{Standardized output generation and downstream interfacing}\label{subsec:karate_output}

The standardized output generation of \textit{Karate Club} ensures that unsupervised learning algorithms which serve the same purpose always return the same type of output with a consistent data point ordering.
There is a very important consequence of this design principle. When a certain type of algorithm is replaced with the same type of algorithm, the downstream code which uses the output of the upstream unsupervised model does not have to be changed. Specifically the outputs generated with our framework use the following data structures:

\begin{itemize}
\item \textit{Node embedding algorithms} (neighbourhood preserving, attributed and structural) always return a \textit{NumPy} float array when the \textit{get\_embedding} method is called. The number of rows in the array is the number of vertices and the row index always corresponds to the vertex index. Furthermore, the number of columns is the number of embedding dimensions. 
\item \textit{Whole graph embedding methods} (spectral fingerprints, implicit matrix factorization techniques) return a \textit{NumPy} float array when the \textit{get\_embedding} method is called. The row index corresponds to the position of a single graph in the list of graphs inputted. In the same way, columns represent the embedding dimensions. 
\item \textit{Community detection procedures} return a dictionary when the \textit{get\_memberships} method is called. Node indices are keys and the values corresponding to the keys are the community memberships of vertices. Certain graph clustering techniques create a node embedding in order to find vertex clusters. These return a \textit{NumPy} float array when the \textit{get\_embedding} method is called. This array is structured like the ones returned by node embedding algorithms. 
\end{itemize}

We demonstrate the standardized output generation and interfacing by the code fragment in Figure \ref{fig:integrated}. We create clusterings of a random graph and return dictionaries containing the cluster memberships. Using the external community library we can calculate the modularity of these clusterings (lines 15-16). This shows that the standardized output generation makes interfacing with external graph mining and machine learning libraries easy.

\subsection{Limitations}\label{subsec:karate_limits}
The current design of \textit{Karate Club} has certain limitations and we make strong assumptions about the input. We assume that that the \textit{NetworkX} graph is undirected and consists of a single strongly connected component. All algorithms assume that nodes are indexed with integers consecutively and the starting node index is 0. Moreover, we assume that the graph is not multipartite, nodes are homogeneous and edges are unweighted (each edge has a unit weight).

In case of the whole graph embedding algorithms \cite{sf, fgsd, graph2vec, gl2vec, netlsd, geoscattering} all graphs in the set of graphs must amend the previously listed requirements with respect to the input. The Weisfeiler-Lehman feature based embedding techniques \cite{graph2vec, gl2vec} allow nodes to have a single string feature which can be accessed with the \textit{feature} key. Without the presence of this key these algorithms default to the use of degree centrality as a node feature. 
\section{Experimental Evaluation}\label{sec:karate_club_experiments}

In the experimental evaluation of \textit{Karate Club} we will demonstrate two things. First, we will show that the implemented algorithms have a good performance with respect to embedding and extracted community quality on a variety of machine learning problems. Second, we support evidence that those algorithms which in theory scale linearly with the input size (number of nodes or number of graphs) scale linearly using our framework in practice. Throughout these experiments we will always use the standard hyperparameter settings of the 1.0. release of our package.
\subsection{Learning performance}
The evaluation of the representation quality focuses on three types of machine learning tasks. These are: community detection with ground truth communities, node classification with the node embeddings, and whole graph classification with graph level embeddings.

\subsubsection{Datasets} In order to evaluate the performance of vertex level algorithms (node embedding and community detection) we used attributed web, collaboration and social networks which are publicly available on \textit{SNAP}  \cite{musae, snap}. We decided to use attributed networks because a large number of algorithms in \textit{Karate Club} can exploit the presence of node features. These datasets are the following:

\begin{itemize}
    \item \textit{Wikipedia Crocodiles:} In this graph nodes represent Wikipedia pages and edges are mutual links. The vertex features describe the presence of nouns in the article and the binary target variable indicates the volume of traffic on the site.   
    \item \textit{GitHub Developers:} Vertices in this network are developers who use GitHub and edges represent mutual follower relationships between the users. Features are derived based on location, biography and other metadata, the binary target variable is whether someone is a machine learning or web developer. 
    \item \textit{Twitch England:} Nodes of this graph are Twitch users from England and edges are mutual friendships between them. Node features were extracted based on the streaming history of the users while the binary node class describes whether the user creates explicit content. 
    \item \textit{Facebook Page-Page:} A network of verified Facebook pages where nodes are pages and the links between nodes are mutual likes. Features are distilled from the page descriptions and the target is the category of the Facebook page (Politicians, Governments, Companies, TV Shows).
\end{itemize}
The descriptive statistics of these node level datasets are summarized in Table \ref{fig:node_level_statistics}. As one can see these networks have a large variety of size, level of clustering and diameter.

\begin{table}[htbp!]

	\centering
	\caption{The social networks used for node level algorithms.}
	\label{fig:node_level_statistics}
	
	\centering{\small
		\begin{tabular}{lcccc}
			& \specialcell{\textbf{Wikipedia}\\\textbf{Crocodiles}} &\specialcell{ \textbf{GitHub}\\\textbf{Developers}} & \specialcell{\textbf{Twitch}\\\textbf{England}}& \specialcell{\textbf{Facebook}\\\textbf{Page-Page}} \\[0.3em]
			\hline
			\textbf{Nodes}          &11,631&37,700&7,126&22,470\\[0.3em]
			\textbf{Density}          &0.003&0.001&0.002&0.001\\[0.3em]
			\textbf{Transitivity}          &0.026&0.013&0.042&0.232\\[0.3em]
			\textbf{Diameter}          &11&7&10&15\\[0.3em]
			\textbf{Features}          &13,183&4,005&2,545&4,714\\[0.3em]
			\hline
	\end{tabular}}
\end{table}

%
		
%

Graph level embedding algorithms were evaluated on a variety of web and social graph datasets which we collected specifically for this paper. We made these graph collections publicly available.\footnote{https://snap.stanford.edu/data/} The graph collections used for predictive performance evaluation are the following:

\begin{itemize}
    \item \textit{Reddit Threads:} Discussion and non-discussion based threads from Reddit which we collected in May 2018. The task is to predict whether a thread is discussion based. 
    \item \textit{Twitch Egos:} The ego-nets of Twitch users who participated in the partnership program in April 2018. The binary classification task is to predict using the ego-net whether the central gamer plays a single or multiple games. 
    \item \textit{Github Stargazers:} The social networks of developers who starred popular machine learning and web development repositories until 2019 August. The task is to decide whether a social network belongs to a web or machine learning repository. 
    \item \textit{Deezer Egos:} The ego-nets of Eastern European users collected from the music streaming service Deezer in February 2020. The related task is the prediction of gender for the ego node in the graph.  
\end{itemize}
\begin{table}[htbp!]
	\centering
	\caption{Statistics of graph datasets used for graph level algorithms.}
	\label{fig:graph_level_statistics}
	
	{\footnotesize
\setlength\tabcolsep{4pt} 

\begin{tabular}{cccccccc}
            &        & \multicolumn{2}{c}{\textbf{Nodes}} & \multicolumn{2}{c}{\textbf{Density}} & \multicolumn{2}{c}{\textbf{Diameter}} \\[0.25em]
            \cline{3-8} 
\textbf{Dataset}     & \textbf{Graphs} & \textbf{Min}         & \textbf{Max}         & \textbf{Min}          & \textbf{Max}          & \textbf{Min}           & \textbf{Max}          \\[0.25em]\hline
\textbf{Reddit Threads} &   203,088     &       11      &         97    & 0.021             &  0.382     & 2               &   27           \\[0.3em]
\textbf{Twitch Egos}   &127,094&14 &52 &0.038  &0.967  &  1             &       2       \\[0.3em]
\textbf{GitHub StarGazers}   & 12,725      &    10        &957      & 0.003     &       0.561            &     2          &18              \\[0.3em]
\textbf{Deezer Egos}   &   9,629     &   11          &     363        &           0.015   &    0.909          &    2           &      2        \\[0.3em]
			\hline
		\end{tabular}}
		
	\end{table}

We listed the size of these datasets with the respective descriptive statistics in Table \ref{fig:graph_level_statistics}. It is worth noting that the \textit{Reddit Threads} and \textit{Twitch Egos} both have at least 10 fold more graphs than the social graph datasets which are widely used for graph classification evaluation \cite{yanardag}.  We would also like to emphasize that the use of graph kernels would not be feasible on graph datasets which are this numerous. 

\begin{table}[htbp!]

	\caption[Mean NMI values with standard errors on the node level datasets calculated from 100 runs.]{Mean NMI values with standard errors on the node level datasets calculated from 100 runs.}
	\label{fig:community_detection_performance}
	\centering{\footnotesize
		\begin{tabular}{lcccc}
			& \specialcell{\textbf{Wikipedia}\\\textbf{Crocodiles}} &\specialcell{ \textbf{GitHub}\\\textbf{Developers}} & \specialcell{\textbf{Twitch}\\\textbf{England}}& \specialcell{\textbf{Facebook}\\\textbf{Page-Page}} \\[0.45em]
			\hline
			\textbf{DANMF}    \cite{danmf}       &$.051\pm.001$&$  .083\pm .001$
			&$.007\pm.001$&$.164\pm.001$\\[0.75em]
			\textbf{M-NMF}     \cite{mnmf}  &$.063\pm.001$&$.084\pm.001$
			&$.004\pm.001$&$.068\pm.001$\\[0.75em]
			\textbf{NNSED}   \cite{nnsed}     &$.063\pm.001$&$.034\pm.001$
			&$.004\pm.001$&$.072\pm.001$\\[0.75em]
			\textbf{SymmNMF}    \cite{symmnmf}   &$.062\pm.001$&$.074\pm.001$
			&$.007\pm.001$&$.206\pm.001$\\[0.75em]
			\textbf{Ego-Splitting} \cite{egosplitting} &$.157\pm.001$&$\textbf{.202}\pm\textbf{.001}$
			&$\textbf{.223}\pm\textbf{.001}$&$.346\pm.001$\\[0.75em]	
			\hline
			\textbf{EdMot}       \cite{edmot}    &$.085\pm.001$&$.180\pm.001$
			&$.008\pm.001$&$.272.\pm.001$\\[0.75em]
			\textbf{LabelProp}    \cite{label_propagation}   &$.119\pm.001$&$.090\pm.002$
			&$.003\pm.001$&$.320\pm.004$\\[0.75em]
			
			\textbf{SCD}  \cite{scd} &$\textbf{.181}\pm\textbf{.001}$&$.189\pm.001$
			&$.169\pm.001$&$\textbf{.386}\pm\textbf{.001}$\\[0.75em]

			\textbf{GEMSEC}\cite{gemsec}   &$.102\pm.001$&$  .127\pm.001$
			&$.008\pm.002$&$.244\pm.001$\\[0.75em]
			\hline \\[-1ex]
	\end{tabular}}
\end{table}

\subsubsection{Community Detection} We evaluate the community detection performance by running the clustering algorithms on the node level datasets. In case of overlapping community detection algorithms \cite{bigclam, danmf, mnmf, nnsed, symmnmf,egosplitting} we assigned each node to the cluster that has the strongest affiliation score with the node (ties were broken randomly). The metric used for the clustering performance measurement is the average normalized mutual information (henceforth NMI) score calculated between the cluster membership vector and the factual class memberships. We report in Table \ref{fig:community_detection_performance} the NMI averages with the standard errors calculated from 100 experimental runs. 

Looking at Table \ref{fig:community_detection_performance} first we notice that the non-overlapping community detection techniques \cite{edmot, label_propagation, scd, gemsec,egosplitting} materially outperform the overlapping models  which create latent spaces \cite{bigclam,danmf, mnmf, nnsed, symmnmf} on every dataset in terms of NMI. Second, those algorithms which create clusters based on the presence of closed triangles (SCD \cite{scd}, Ego-Splitting \cite{egosplitting}) have a general strong performance. Finally, on problems where  it can be assumed that the class membership vector is associated with structural properties (e.g. Wikipedia Crocodiles), the overlapping latent space creating community detection methods perform poorly in terms of NMI.

\subsubsection{Graph classification} In each dataset we created representations for the graphs and use those as predictors for the downstream classification task. We repeated the feature distillation and supervised model training 100 times, used 80\% of graphs for training and 20\% for testing with seeded splits. Using the graph class vectors of the test set and class probabilities outputted by the logistic regression classifier we calculated mean area under the curve (henceforth AUC) values which are presented in Table  \ref{fig:graph_classification_performance} along with their standard errors.  

\begin{table}[h!]
	
	\caption[Mean AUC values with standard errors on the graph level datasets calculated from 100 seed train-test splits.]{Mean AUC values with standard errors on the graph level datasets calculated from 100 seed train-test splits.}
	\label{fig:graph_classification_performance}
	\centering{\footnotesize
		\begin{tabular}{lcccc}
			& \specialcell{\textbf{Reddit}\\\textbf{Threads}} & \specialcell{\textbf{Twitch}\\\textbf{Egos}} & \specialcell{\textbf{GitHub}\\\textbf{StarGazers}}& \specialcell{\textbf{Deezer}\\\textbf{Egos}} \\[0.45em]
			\hline
			\textbf{GL2Vec} \cite{gl2vec}           &$.753\pm.002$&$ .664\pm.002$
			&$.551\pm.001$&$.504\pm.001$\\[0.75em]
			\textbf{Graph2Vec}   \cite{graph2vec}         &$.804\pm.002$&$.702\pm.003$
			&$.585\pm.001$&$.512\pm.001$\\[0.75em]
			\hline
			\textbf{SF}    \cite{sf}        &$.814\pm.002$&$.678\pm.003$
			&$.558\pm.001$&$.501\pm.001$\\[0.75em]
			\textbf{NetLSD} \cite{netlsd}            &$\textbf{.827}\pm\textbf{.001}$&$.631\pm.002$
			&$.632\pm.001$&$.522\pm.001$\\[0.75em]
			\textbf{FGSD}    \cite{fgsd}        &$.825\pm.002$&$.705\pm.003$
			&$.656\pm .001$&$.526\pm.001$\\[0.75em]
		\textbf{GeoScattering}    \cite{geoscattering}        &$.800        \pm.001$&$.697\pm.001$
			&$.546\pm .003$&$.522\pm.003$\\[0.75em]
		\textbf{FEATHER}    \cite{feather}        &$\textbf{.830}\pm\textbf{.002}$&$\textbf{.720}\pm\textbf{.003}$
			&$\textbf{.748}\pm\textbf{.002}$&$\textbf{.540}\pm\textbf{.001}$\\[0.75em]
			\hline \\[-1ex]
		\end{tabular}}
	\end{table}

The results presented in Table \ref{fig:graph_classification_performance} show that the representations created by implicit factorization  \cite{graph2vec, gl2vec} and spectral finger printing \cite{sf,netlsd, fgsd, feather} techniques are predictive on most problems. In addition, we see evidence that algorithms from the latter group create somewhat higher quality representations.

\subsubsection{Node classification} In this series of experiments we evaluated the node classification performance on the node level datasets. For each graph we learned a node embedding and used the features of this node embedding as predictors for a downstream logistic (softmax) regression model. We repeated the embedding and supervised model training 100 times, used 80\% of the nodes for training and 20\% for testing with seeded splits. Using the target vectors of the test set and the class probabilities outputted by the downstream model we calculated mean AUC scores. These average AUC values are reported in Table  \ref{fig:node_classification_performance} with standard errors. 
The results in Table \ref{fig:node_classification_performance} generally demonstrate that the included neighbourhood based \cite{deepwalk, walklets, diff2vec, grarep, netmf, boostne,  nmfadmm, hope, eigenmap}, structural role preserving \cite{role2vec, graphwave}, and attributed \cite{musae, bane, tene, tadw, fscnmf, sine} node embedding techniques all generate reasonable quality representations for this classification task. There are additional conclusions; (i) multi-scale node embeddings such as \textit{GraRep} \cite{grarep}, \textit{Walklets}, \cite{walklets}, and \textit{MUSAE} \cite{musae} create high quality node features , (ii) combining neighbourhood and attribute information results in the best representations \cite{musae,sine}, (iii) there is not a single model which is generally superior.

\subsection{Scalability}
We perform scalability tests for all three types of algorithms (community detection, node and whole graph embedding). For each of these categories we investigate the scalability of 4 chosen algorithms. We use Erdos-Renyi graphs where the input size and graph density can be manipulated directly.

\begin{table}[h!]
	
	\caption[Mean AUC values with standard errors on the node level datasets calculated from 100 seed train-test splits.]{Mean AUC values with standard errors on the node level datasets calculated from 100 seed train-test splits.}
	\label{fig:node_classification_performance}
	\centering{\footnotesize
		\begin{tabular}{lcccc}
			& \specialcell{\textbf{Wikipedia}\\\textbf{Crocodiles}} &\specialcell{ \textbf{GitHub}\\\textbf{Developers}} & \specialcell{\textbf{Twitch}\\\textbf{England}}& \specialcell{\textbf{Facebook}\\\textbf{Page-Page}} \\[0.45em]
			\hline
			\textbf{BoostNE}   \cite{boostne}         &$.685\pm.001$&$.845\pm.001$
			&$.576\pm.001$&$.752\pm.001$\\[0.75em]
	\textbf{NodeSketch} \cite{nodesketch}            &$.722\pm.001$&$.631\pm.001$
			&$.520\pm.001$&$.579\pm.001$\\[0.75em]		\textbf{Diff2Vec}  \cite{diff2vec}          &$.832\pm.001$&$.858\pm.001$
			&$.589\pm.001$&$.873\pm.001$\\[0.75em]
			\textbf{NetMF}    \cite{netmf}        &$.866\pm.001$&$.867\pm.001$
			&$.629\pm.002$&$.946\pm.001$\\[0.75em]
			\textbf{Walklets}   \cite{walklets}         &$.875\pm.001$&$.899\pm.002$
			&$.622\pm.001$&$.973\pm.001$\\[0.75em]
			\textbf{HOPE} \cite{hope}            &$.870\pm.001$&$.844\pm.001$
			&$.612\pm.001$&$.909\pm.001$\\[0.75em]
			\textbf{GraRep}   \cite{grarep}         &$.888\pm.002$&$.876\pm.001$
	&$.609\pm.001$&$.952\pm.001$\\[0.75em]
			\textbf{DeepWalk}   \cite{deepwalk}         &$.850\pm.001$&$.872\pm.002$
			&$.597\pm.002$&$.877\pm.001$\\[0.75em]
			\textbf{NMF-ADMM}  \cite{nmfadmm}          &$.747\pm.001$&$.784\pm.001$
			&$.619\pm.001$&$.937\pm.001$\\[0.75em]
			\textbf{LAP} \cite{eigenmap}            &$.784\pm.001$&$.529\pm.001$
			&$.511\pm.001$&$.501\pm.001$\\[0.75em]			
			\hline
			\textbf{GraphWave}  \cite{graphwave}          &$.517\pm.001$&$.620\pm.001$
			&$.583\pm.001$&$.613\pm.001$\\[0.75em]
			\textbf{Role2Vec}  \cite{role2vec}          &$.845\pm.001$&$.862\pm.002$
			&$.601\pm.002$&$.903\pm.002$\\[0.75em]	\hline		
			\textbf{BANE}     \cite{bane}       &$.866\pm.002$&$.570\pm.001$
			&$.551\pm.001$&$.970\pm.002$\\[0.75em]
			\textbf{TENE}     \cite{tene}       &$.907\pm.001$&$.874\pm.001$
			&$.615\pm.001$&$.886\pm.001$\\[0.75em]
			\textbf{TADW}       \cite{tadw}     &$.896\pm.001$&$.817\pm.001$
			&$.612\pm.002$&$.871\pm.001$\\[0.75em]
			\textbf{FSCNMF}      \cite{fscnmf}      &$.912\pm.001$&$.856\pm.002$
			&$.621\pm.001$&$.891\pm.001$\\[0.75em]
			
			\textbf{SINE}      \cite{sine}      &$.904\pm.001$&$ \textbf{.910}\pm\textbf{.002}$
			&$\textbf{.646}\pm \textbf{.001}$&$.979\pm.001$\\[0.75em]			
			
			\textbf{MUSAE}      \cite{musae}      &$\textbf{.931}\pm\textbf{.001}$&$.903\pm.001$
			&$.628\pm.001$&$\textbf{.981}\pm\textbf{.001}$\\[0.75em]
			\hline \\[-1ex]
		\end{tabular}}
	\end{table}

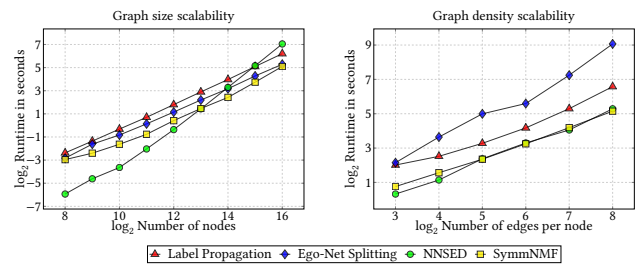
\begin{figure}[ht!]
\centering
\begin{tikzpicture}[scale=0.24,transform shape]
\tikzset{font={\fontsize{22pt}{12}\selectfont}}
\begin{groupplot}[group style={group size=2 by 1,
		          horizontal sep=110pt,
		          vertical sep=70pt,ylabels at=edge left},
	              width=0.9\textwidth,
	              height=0.65\textwidth,
	              grid=major,
	              grid style={dashed, gray!40},
	              scaled ticks=false,
	              inner axis line style={-stealth}]
 \nextgroupplot[ytick={-7,-5,-3,-1,1,3,5,7,9,11},
	              xtick={6,8,10,12,14,16},
	xlabel=$\log_2$ Number of nodes,
	ylabel=$\log_2$ Runtime in seconds,
	enlargelimits=0.1,
	legend style = { column sep = 10pt, legend columns = -1, legend to name = grouplegend, title = Graph size scalability}]
	
	\addplot[mark=triangle*,opacity=0.8,mark options={black,fill=red},mark size=7pt]
	coordinates {
(8,-2.374)
(9,-1.359)
(10,-0.318)
(11,0.702)
(12,1.798)
(13,2.896)
(14,3.968)
(15,5.093)
(16,6.202)
	};\addlegendentry{Label Propagation}%
	\addplot[mark=diamond*,opacity=0.8,mark options={black,fill=blue},mark size=7pt]
	coordinates {
(8,-2.832)
(9,-1.621)
(10,-0.83)
(11,0.133)
(12,1.153)
(13,2.197)
(14,3.171)
(15,4.276)
(16,5.287)
	};\addlegendentry{Ego-Net Splitting}%
	\addplot[mark=*,opacity=0.8,mark options={black,fill=green},mark size=5pt]
	coordinates {

(8,-5.934)
(9,-4.616)
(10,-3.641)
(11,-2.035)
(12,-0.365)
(13,1.433)
(14,3.291)
(15,5.160)
(16,7.047)

	};\addlegendentry{NNSED}%
	
	\addplot[mark=square*,opacity=0.8,mark options={black,fill=yellow},mark size=5pt]
	coordinates {
(8,-2.958)
(9,-2.402)
(10,-1.63)
(11,-0.769)
(12,0.414)
(13,1.46)
(14,2.422)
(15,3.74)
(16,5.099)
	};\addlegendentry{SymmNMF}%

	\nextgroupplot[ytick={1,3,5,7,9,11},
	xtick={3,4,5,6,7,8},
	xlabel=$\log_2$ Number of edges per node,
	ylabel=$\log_2$ Runtime in seconds,
	enlargelimits=0.1,
	legend style = { column sep = 10pt, legend columns = -1, legend to name = grouplegend, title = Graph density scalability}]
	
	\addplot[mark=triangle*,opacity=0.8,mark options={black,fill=red},mark size=7pt]
	coordinates {
(3,2.012)
(4,2.52)
(5,3.277)
(6,4.173)
(7,5.295)
(8,6.578)

	};\addlegendentry{Label Propagation}%
	\addplot[mark=diamond*,opacity=0.8,mark options={black,fill=blue},mark size=7pt]
	coordinates {
(3,2.149)
(4,3.642)
(5,4.994)
(6,5.594)
(7,7.245)
(8,9.065)

	};\addlegendentry{Ego-Net Splitting}%
	\addplot[mark=*,opacity=0.8,mark options={black,fill=green},mark size=5pt]
	coordinates {
(3,0.324)
(4,1.14)
(5,2.375)
(6,3.297)
(7,4.054)
(8,5.286)
	};\addlegendentry{NNSED}%
	
	\addplot[mark=square*,opacity=0.8,mark options={black,fill=yellow},mark size=5pt]
	coordinates {
(3,0.759)
(4,1.565)
(5,2.35)
(6,3.248)
(7,4.19)
(8,5.149)

	};\addlegendentry{SymmNMF}%

	\end{groupplot}	
	\node at ($(group c1r1) + (10.0cm,-7.5cm)$) {\ref{grouplegend}}; 
	\end{tikzpicture}
\caption{Scalability of the community detection procedures in Karate Club. We vary the number of nodes and the density of an Erdos-Renyi graph.}\label{fig:community_scaling}
\end{figure}

Figure \ref{fig:community_scaling} plots runtime against size and density of the clustered while the average number of edges is fixed to be 10. In the densification scenario we clustered a graph with $2^{12}$ nodes. Non-overlapping community detection techniques show a
remarkable scalability with respect to graph size increase, and we also see that the densification of the graph results in longer runtimes.

\begin{figure}[ht!]
\centering
\begin{tikzpicture}[scale=0.24,transform shape]
\tikzset{font={\fontsize{22pt}{12}\selectfont}}
\begin{groupplot}[group style={group size=2 by 1,
		          horizontal sep=110pt,
		          vertical sep=70pt,ylabels at=edge left},
	              width=0.9\textwidth,
	              height=0.65\textwidth,
	              grid=major,
	              grid style={dashed, gray!40},
	              scaled ticks=false,
	              inner axis line style={-stealth}]
	          \nextgroupplot[ytick={-5,-3,-1,1,3,5,7,9,11},
	              xtick={8,10,12,14,16},
	xlabel=$\log_2$ Number of nodes,
	ylabel=$\log_2$ Runtime in seconds,
	enlargelimits=0.1,
	legend style = { column sep = 10pt, legend columns = -1, legend to name = grouplegend, title = Graph size scalability}]
	
	\addplot[mark=triangle*,opacity=0.8,mark options={black,fill=red},mark size=7pt]
	coordinates {
(8,0.194)
(9,1.297)
(10,2.315)
(11,3.324)
(12,4.354)
(13,5.425)
(14,6.521)
(15,7.547)
(16,8.543)

	};\addlegendentry{DeepWalk}%
	\addplot[mark=diamond*,opacity=0.8,mark options={black,fill=blue},mark size=7pt]
	coordinates {
(8,0.926)
(9,1.987)
(10,2.973)
(11,4.02)
(12,5.068)
(13,6.077)
(14,7.159)
(15,8.208)
(16,9.439)

	};\addlegendentry{Walklets}%
	\addplot[mark=*,opacity=0.8,mark options={black,fill=green},mark size=5pt]
	coordinates {
(8,-4.5)
(9,-4.054)
(10,-3.761)
(11,-2.577)
(12,-1.513)
(13,-0.465)
(14,0.524)
(15,1.594)
(16,2.623)
	};\addlegendentry{NetMF}%
	
	\addplot[mark=square*,opacity=0.8,mark options={black,fill=yellow},mark size=5pt]
	coordinates {
(8,-2.076)
(9,-1.241)
(10,-0.389)
(11,0.572)
(12,1.647)
(13,2.673)
(14,3.681)
(15,4.684)
(16,5.702)
	};\addlegendentry{BoostNE}%

	\nextgroupplot[ytick={-1,1,3,5,7,9},
	xtick={3,4,5,6,7,8},
	xlabel=$\log_2$ Number of edges per node,
	ylabel=$\log_2$ Runtime in seconds,
	enlargelimits=0.1,
	legend style = { column sep = 10pt, legend columns = -1, legend to name = grouplegend, title = Graph density scalability}]
	
	\addplot[mark=triangle*,opacity=0.8,mark options={black,fill=red},mark size=7pt]
	coordinates {
(3,4.368)
(4,4.508)
(5,4.807)
(6,5.181)
(7,5.635)
(8,6.286)

	};\addlegendentry{DeepWalk}%
	\addplot[mark=diamond*,opacity=0.8,mark options={black,fill=blue},mark size=7pt]
	coordinates {
(3,5.093)
(4,5.163)
(5,5.363)
(6,5.614)
(7,5.987)
(8,6.521)

	};\addlegendentry{Walklets}%
	\addplot[mark=*,opacity=0.8,mark options={black,fill=green},mark size=5pt]
	coordinates {
(3,-0.382)
(4,1.309)
(5,2.691)
(6,3.357)
(7,3.574)
(8,3.986)

	};\addlegendentry{NetMF}%
	
	\addplot[mark=square*,opacity=0.8,mark options={black,fill=yellow},mark size=5pt]
	coordinates {
(3,2.819)
(4,4.583)
(5,5.107)
(6,5.637)
(7,5.972)
(8,6.231)
	};\addlegendentry{BoostNE}%

	\end{groupplot}	
	\node at ($(group c1r1) + (10.0cm,-7.5cm)$) {\ref{grouplegend}}; 
	\end{tikzpicture}
\caption{Scalability of node embedding procedures in Karate Club. We vary the number of nodes and the density of an Erdos-Renyi graph.}\label{fig:node_embedding_scaling}
\end{figure}
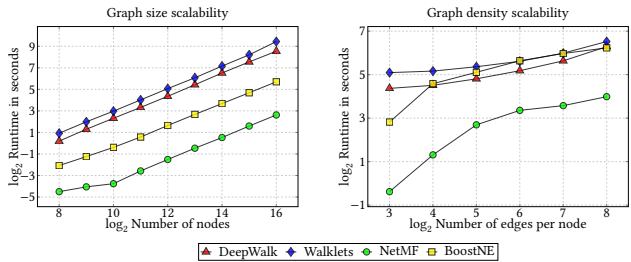

We measured the same way how the average runtime of node embedding varies with input size changes and densification and plotted these in Figure \ref{fig:node_embedding_scaling}. These results show that under no preferential attachment all of the included methods scale linearly with input size changes. Moreover, implicit factorization runtimes are unaffected by the densification of the graph.

\begin{figure}[ht!]
\centering
\begin{tikzpicture}[scale=0.24,transform shape]
\tikzset{font={\fontsize{22pt}{12}\selectfont}}
\begin{groupplot}[group style={group size=2 by 1,
		          horizontal sep=110pt,
		          vertical sep=70pt,ylabels at=edge left},
	              width=0.9\textwidth,
	              height=0.65\textwidth,
	              grid=major,
	              grid style={dashed, gray!40},
	              scaled ticks=false,
	              inner axis line style={-stealth}]
	              \nextgroupplot[ytick={-3,-1,1,3,5,7,9,11},
	              xtick={6,8,10,12,14,16},
	xlabel=$\log_2$ Number of graphs,
	ylabel=$\log_2$ Runtime in seconds,
	enlargelimits=0.1,
	legend style = { column sep = 10pt, legend columns = -1, legend to name = grouplegend, title = Graph count scalability}]
	
	\addplot[mark=triangle*,opacity=0.8,mark options={black,fill=red},mark size=7pt]
	coordinates {
(6,-0.459)
(7,0.49)
(8,1.374)
(9,2.272)
(10,3.173)
(11,4.191)
(12,5.158)
(13,6.151)
(14,7.106)
(15,8.057)
(16,9.083)
	};\addlegendentry{Graph2Vec}%
	\addplot[mark=diamond*,opacity=0.8,mark options={black,fill=blue},mark size=7pt]
	coordinates {
(6,-2.984)
(7,-2.026)
(8,-1.044)
(9,-0.026)
(10,0.964)
(11,2.031)
(12,3.055)
(13,4.057)
(14,5.083)
(15,6.056)
(16,7.076)

	};\addlegendentry{FGSD}%
	\addplot[mark=*,opacity=0.8,mark options={black,fill=green},mark size=5pt]
	coordinates {
(6,-2.727)
(7,-1.751)
(8,-0.754)
(9,0.257)
(10,1.266)
(11,2.264)
(12,3.272)
(13,4.274)
(14,5.3)
(15,6.343)
(16,7.305)
	};\addlegendentry{SF}%
	
	\addplot[mark=square*,opacity=0.8,mark options={black,fill=yellow},mark size=5pt]
	coordinates {
(6,1.184)
(7,2.133)
(8,3.125)
(9,4.07)
(10,5.061)
(11,5.948)
(12,6.893)
(13,7.839)
(14,8.844)
(15,9.72)
(16,10.697)
	};\addlegendentry{GL2Vec}%

	\nextgroupplot[ytick={1,3,5,7,9,11},
	xtick={3,5,7,9,11},
	xlabel=$\log_2$ Number of nodes,
	ylabel=$\log_2$ Runtime in seconds,
	enlargelimits=0.1,
	legend style = { column sep = 10pt, legend columns = -1, legend to name = grouplegend, title = Graph size scalability}]
	
	\addplot[mark=triangle*,opacity=0.8,mark options={black,fill=red},mark size=7pt]
	coordinates {
(3,-0.122)
(4,2.269)
(5,3.209)
(6,4.163)
(7,5.093)
(8,6.023)
(9,6.961)
(10,7.972)
(11,8.775)
	};\addlegendentry{Graph2Vec}%
	\addplot[mark=diamond*,opacity=0.8,mark options={black,fill=blue},mark size=7pt]
	coordinates {
(3,0.603)
(4,0.71)
(5,0.96)
(6,2.004)
(7,3.009)
(8,4.845)
(9,6.914)
(10,9.126)
(11,12.184)
	};\addlegendentry{FGSD}%
	\addplot[mark=*,opacity=0.8,mark options={black,fill=green},mark size=5pt]
	coordinates {
(3,0.663)
(4,0.889)
(5,1.316)
(6,1.95)
(7,3.479)
(8,4.494)
(9,6.459)
(10,8.243)
(11,10.352)
	};\addlegendentry{SF}%
	
	\addplot[mark=square*,opacity=0.8,mark options={black,fill=yellow},mark size=5pt]
	coordinates {
(3,1.208)
(4,3.995)
(5,4.97)
(6,6.012)
(7,6.984)
(8,8.0)
(9,8.974)
(10,10.009)
(11,10.76)
	};\addlegendentry{GL2Vec}%

	\end{groupplot}	
	\node at ($(group c1r1) + (10.0cm,-7.5cm)$) {\ref{grouplegend}}; 
	\end{tikzpicture}
\caption{Scalability of graph embedding and summarization procedures in Karate Club. We vary the number of Erdos-Renyi graphs and their size.}\label{fig:graph_embedding_scaling}
\end{figure}
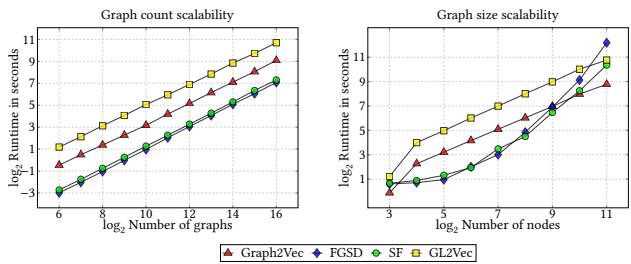

In case of the whole graph representation we plotted the average runtime as a function of the number of graphs and their size on Figure \ref{fig:graph_embedding_scaling}. The base graph used for the first plot had 64 nodes and 5 edges per node and for the second plot we used $2^{10}$ graphs. First, a takeaway is that the runtime increases linearly with the size of the dataset assuming that size of the graphs is homogeneous. Second, the spectral fingerprinting techniques \cite{sf, fgsd} do not scale well when the size of the graphs is increased which was expected.

\section{Related Work}\label{sec:karate_club_related_work}

In this section we discuss how the design of our framework is related to existing machine learning frameworks, what differentiates it from other graph mining tools.

\subsection{API oriented machine learning frameworks}
\textit{Scikit-learn}~\cite{scikit,scikitapi} is a machine learning framework with consistent and easy to use design. The \textit{scikit-learn} models are characterised by models with a consistent API, their constructors have encapsulated sensible hyperparameters and utilize widely used Python data structures for data ingestion and output generation. 
This compositional design of the framework results in a low number of model classes, reusable model blocks and enables fast deployment. The \textit{Karate Club} API draws heavily from the ideas of \textit{scikit-learn} and the output generated by \textit{Karate Club} is suitable as input for \textit{scikit-learn}'s machine learning procedures.
\subsection{Graph mining libraries}
The \textit{Karate Club} framework is differentiated from other graph mining libraries because of lightweight prerequisites and wide coverage of the learning techniques. First, the \textit{SNAP} and \textit{GraphTool} packages both have C++ prerequisites which have to be pre-compiled and installed. Our framework only has Python dependencies and builds on top of the \textit{NetworkX} project. Second, the \textit{SNAP} \cite{snap} library only covers specific methods which were created by the authors of the framework. The \textit{NetworkX} \cite{networkx} and \textit{GraphTool} \cite{graphtool} libraries only provide community detection tools. Node and whole graph embedding is not supported by these frameworks. 


\section{Conclusion and Future Directions}\label{sec:karate_club_conclusion}
In this work we described \textit{Karate Club} a Python framework built on the open source packages \textit{NetworkX} \cite{networkx}, \textit{PyGSP} \cite{pygsp}, \textit{Gensim} \cite{gensim},  \textit{NumPy} \cite{numpy}, and \textit{SciPy Sparse} \cite{scipy} which performs unsupervised learning on graph data. Specifically, it supports community detection, node embedding, and whole graph embedding techniques.

We discussed in detail the design principles which we followed when we created \textit{Karate Club}, standard hyperparameter encapsulation, the assumptions about the format of input data and generated output, and available public methods. In order to demonstrate these principles we included illustrative examples of code. In a series of experiments on real world datasets we validated that the machine learning models in \textit{Karate Club} produce high quality clusters and embeddings. We also demonstrated on synthetic data that the linear runtime algorithms scale well with increasing input size. 

As discussed, \textit{Karate Club} has certain limitations with regards to the types of graphs that it can handle. In the future we plan to extend it to operate on directed and weighted graphs. Another aim is to provide a general framework for unsupervised learning algorithms on heterogeneous, multiplex, temporal graphs and procedures for the hyperbolic embedding of nodes~\cite{sarkar2011low,verbeek2014metric}.

\section*{Acknowledgements}
Benedek Rozemberczki was supported by the Centre for Doctoral Training in Data Science, funded by EPSRC (grant EP/L016427/1).

\bibliographystyle{ACM-Reference-Format}

\bibliography{main}

\end{document}